\documentclass[10pt,twocolumn,letterpaper]{article}

\usepackage[pagenumbers]{cvpr} 

\usepackage[dvipsnames]{xcolor}

\definecolor{cvprblue}{rgb}{0.21,0.49,0.74}
\usepackage[pagebackref,breaklinks,colorlinks,citecolor=cvprblue]
{hyperref}
\usepackage{cite}
\usepackage{amsmath,amssymb,amsfonts}
\usepackage{algorithmic}
\usepackage{graphicx}
\usepackage{textcomp}
\usepackage{xcolor}
\usepackage{gensymb}
\usepackage{amssymb}
\usepackage{hyperref}
\usepackage{tabularray}
\usepackage{tabularx}
\usepackage{threeparttable}
\usepackage{caption}
\usepackage{subcaption}
\usepackage{mathtools}
\usepackage{multirow}
\usepackage{textcomp}
\usepackage{stfloats}
\newcommand{\RR}{\mathbb{R}}
\newcommand{\codebook}{Z}

\newcommand{\decoder}{G}
\newcommand{\encoder}{E}
\newcommand{\quantize}{Q}
\newcommand{\quantizedcode}{\mathbf{z}_{q}}
\newcommand{\rangeimage}{\mathbf{x}}
\newcommand{\mask}{\mathbf{x}_{m}}
\newcommand{\masklogit}{\mathbf{\hat{x}_{\pi}}}
\newcommand{\reconcleanimage}{\mathbf{\hat{x}}_{r}}
\newcommand{\reconrangeimage}{\mathbf{\hat{x}}}

\newcommand{\expec}{\mathbb{E}_{\rangeimage}}
\newcommand{\reconmask}{\mathbf{\hat{x}}_{m}}
\newcommand{\sigmoid}{\text{sigmoid}}
\newcommand{\trans}{\mathbf{\mathcal{F}}}
\newcommand{\unquantizedcode}{\mathbf{\hat{z}}}
\newcommand{\codebookdim}{n_z}
\newcommand{\height}{\text{H}}
\newcommand{\width}{\text{W}}
\newcommand{\codeindice}{\mathbf{s}}

\newcommand{\blfootnote}[1]{%
  \begingroup
  \renewcommand\thefootnote{}\footnote{#1}%
  \addtocounter{footnote}{-1}%
  \endgroup
  }
\def\paperID{*****} 
\def\confName{CVPR}
\def\confYear{2024}

\title{Taming Transformers for Realistic Lidar Point Cloud Generation}

\author{Hamed Haghighi$^1$ \hspace{-0.4cm}
\and
Amir Samadi$^1$ \hspace{-0.4cm}
\and 
Mehrdad Dianati$^2$ \hspace{-0.4cm}
\and
Valentina Donzella$^1$ \hspace{-0.4cm}
\and
Kurt Debattista$^1$
}

\begin{document}
\maketitle
\begin{abstract}
Diffusion Models (DMs) have achieved State-Of-The-Art (SOTA) results in the Lidar point cloud generation task, benefiting from their stable training and iterative refinement during sampling. However, DMs often fail to realistically model Lidar raydrop noise due to their inherent denoising process. To retain the strength of iterative sampling while enhancing the generation of raydrop noise, we introduce LidarGRIT, a generative model that uses auto-regressive transformers to iteratively sample the range images in the latent space rather than image space. Furthermore, LidarGRIT utilises VQ-VAE to separately decode range images and raydrop masks. Our results show that LidarGRIT achieves superior performance compared to SOTA models on KITTI-360 and KITTI odometry datasets. Code available at:\href{https://github.com/hamedhaghighi/LidarGRIT}{https://github.com/hamedhaghighi/LidarGRIT}.
\end{abstract}
\blfootnote{$^{1}$H. Haghighi, A.Samadi, K. Debattista and V. Donzella are with WMG, University of Warwick, Coventry, U.K. (Corresponding author: \tt Hamed.Haghighi@warwick.ac.uk)}
\blfootnote{$^2$M. Dianati is with the School of Electronics, Electrical Engineering and Computer Science at Queen’s University of Belfast and WMG at the University of Warwick}
\section{Introduction} \label{introdcution}
Light detection and ranging (Lidar) is a critical sensor in autonomous vehicles, providing highly precise 3D environmental scanning. However, realistic simulation of the Lidar sensor poses challenges, involving cumbersome tasks such as creating 3D object models along with running computationally demanding physics-based algorithms. As an alternative, data-driven simulation models, particularly deep generative models have gained traction owing to their exceptional capacity to model high-dimensional data. Initially proposed for generating photo-realistic RGB images, deep generative models have been adapted for Lidar point cloud generation, progressing from early GAN-based frameworks~\cite{Caccia2018DeepGM} to the best-performing diffusion models~\cite{10.1007/978-3-031-20050-2_2}. \par
\begin{figure}[t!]
    \centering
    \begin{subfigure}{\linewidth}
        \centering
        \includegraphics[width=1.0\linewidth]{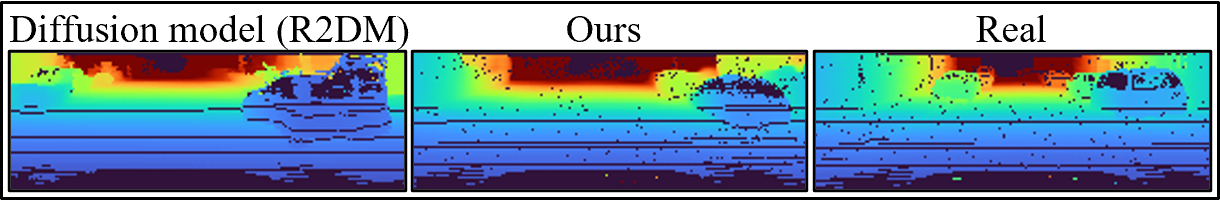}\\{\vspace{-0.15cm}\footnotesize (a)\vspace{+0.05cm}}
        \label{fig:t0}
    \end{subfigure}
    \begin{subfigure}{\linewidth}
        \centering
        \includegraphics[width=\linewidth]{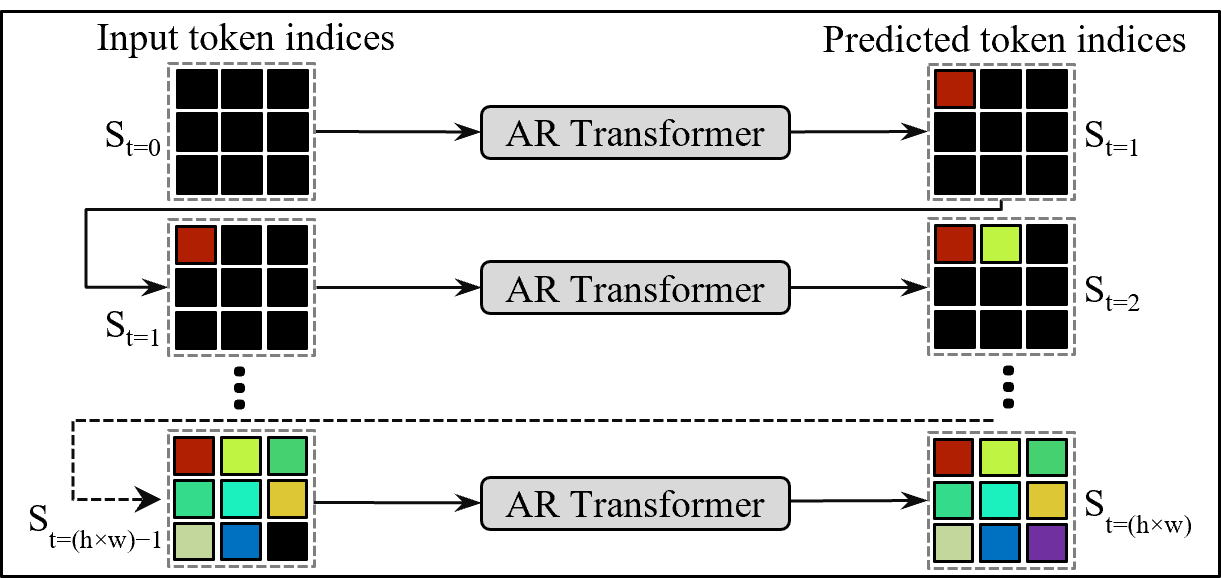}\\{\vspace{-0.15cm}\footnotesize (b)\vspace{+0.05cm}}
        \label{fig:t1}
    \end{subfigure}
    \begin{subfigure}{\linewidth}
    \centering
    \includegraphics[width=\linewidth]{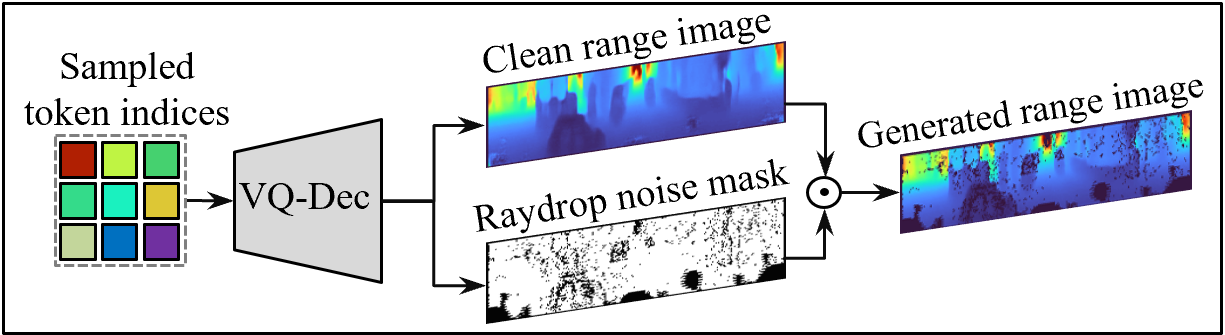}\\{\vspace{-0.15cm}\footnotesize (c)\vspace{-0.2cm}}
    \label{fig:t2}
    \end{subfigure}
    \caption{(a) The range image generated by diffusion model (R2DM~\cite{nakashima2024lidar}) exhibits less realistic raydrop noise compared to our provided sample and the real one. (b) we propose to sample range image in the latent space via Auto-Regressive (AR) transformer~\cite{10.5555/3295222.3295349}. (c) We then generate the raydrop mask and clean range image separately in the image space via VQ-VAE~\cite{DBLP:conf/cvpr/EsserRO21} decoder.}
    \label{fig:teaser}
\vspace{-0.5cm}
\end{figure}

 
Diffusion models (DMs) for Lidar point cloud generation excel mainly due to their stable training and iterative refinement during the sampling process. While they demonstrate proficiency in capturing the 3D shape of point clouds, they face challenges in generating realistic Lidar raydrop noise, resulting in range images that appear unrealistic (refer to Figure~\ref{fig:teaser}\textcolor{red}{a}). This issue arises from the inherent denoising nature of DMs.

We introduce a novel Lidar Generative Range Image Transformer (LidarGRIT) model to incorporate both progressive generation and accurate raydrop noise synthesis.  Our LidarGRIT works with the range image representation of Lidar point cloud, chosen for efficient processing and compatibility with image generative models. The generation process of the LidarGRIT consists of an iterative sampling in the latent space via Auto-Regressive (AR) transformer~\cite{10.5555/3295222.3295349} (refer to Figure~\ref{fig:teaser}\textcolor{red}{b}), and decoding the sampled tokens to range images using an adapted Vector Quantised Variational Auto-Encoder (VQ-VAE~\cite{DBLP:conf/cvpr/EsserRO21}) model (refer to Figure~\ref{fig:teaser}\textcolor{red}{c}). We disentangle the generation of the range image from the raydrop noise mask in the VQ-VAE decoder, inspired by Dusty~\cite{Nakashima2021LearningTD}, and use a separate loss function to reconstruct clean range images and raydrop masks during training. Furthermore, we realised that large VQ-VAE models, primarily designed for high-resolution RGB images, tend to overfit when applied to relatively low-resolution range images. To address this, we propose geometric preservation, aiming to encourage the VQ-VAE to capture input geometry and provide more expressive latent tokens. We compare our LidarGRIT model to SOTA models on KITTI-360 and KITTI odometry datasets. Our model outperforms on nearly all metrics, specifically excelling in the image-based metrics, SWD~\cite{Karras2017ProgressiveGO}. The contributions of this paper can be summarised as follows:
\begin{itemize}
    \item Introduction of LidarGRIT, a novel Lidar point cloud generative model consisting of a two-step generation process: iterative token index sampling through an AR transformer and single-pass range image decoding via an adapted VQ-VAE.
    \item Proposal of two novel techniques to enhance the generation quality: incorporating a separate raydrop estimation loss and enforcing geometry perseverance to increase VQ-VAE generalisability.
    \item Comprehensive evaluation of our LidarGRIT generation by comparing it with SOTA generative models using KITTI-360 and KITTI odometry datasets.
\end{itemize} 
\section{Related-Work} \label{related-work} 
 Caccia~{\it et al.}~\cite{Caccia2018DeepGM} were among the first researchers to apply deep generative models to Lidar point clouds. They converted Lidar point clouds into range images and adapted the Deep Convolutional GAN (DCGAN)~\cite{Radford2015UnsupervisedRL} for point cloud generation. Building on this, Dusty~\cite{Nakashima2021LearningTD,nakashima2022generative} was proposed, which integrates raydrop synthesis into the GAN training process. 
 Another notable model, UltraLiDAR~\cite{xiong2023learning}, adopts a VQ-VAE framework to learn a discrete and compact Lidar representation for point cloud restoration and generation. With the recent achievement of DMs, LidarGen~\cite{10.1007/978-3-031-20050-2_2} and R2DM~\cite{nakashima2024lidar} models were proposed, relying on the score-based and denoising DMs frameworks, respectively. \par Our approach draws inspiration from Dusty~\cite{Nakashima2021LearningTD} framework in the disentanglement of raydrop and range image generation, however, we differentiate by employing a non-adversarial and more stable training using VQ-VAE~\cite{DBLP:conf/cvpr/EsserRO21}, treating the raydrop estimation as a binary classification problem. Our LidarGRIT shares similarities with UltraLiDAR~\cite{xiong2023learning} in its two-stage sampling using VQ-VAE and AR transformer. However, rather than using voxelised Birds-Eye-View (BEV), we represent point clouds with range images that provide a lossless, more compact, and more computationally efficient format. Moreover, we focus on the raydrop noise generation and assess the point cloud generation on both image and point cloud representation.

\begin{figure*}[t!]
    \centering
    \begin{subfigure}{0.6\linewidth}
        \centering
        \includegraphics[width=\linewidth, height=2.95cm]{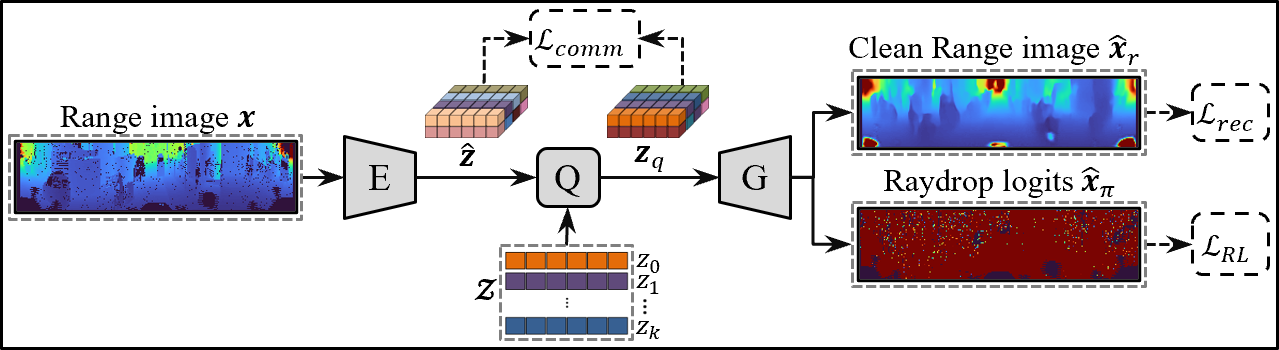}\\{\vspace{-0.13cm}\footnotesize (a) Adapted VQ-VAE model}
        \label{fig:method0}
    \end{subfigure}
    \begin{subfigure}{0.39\linewidth}
        \centering
        \includegraphics[width=\linewidth, height=2.95cm]{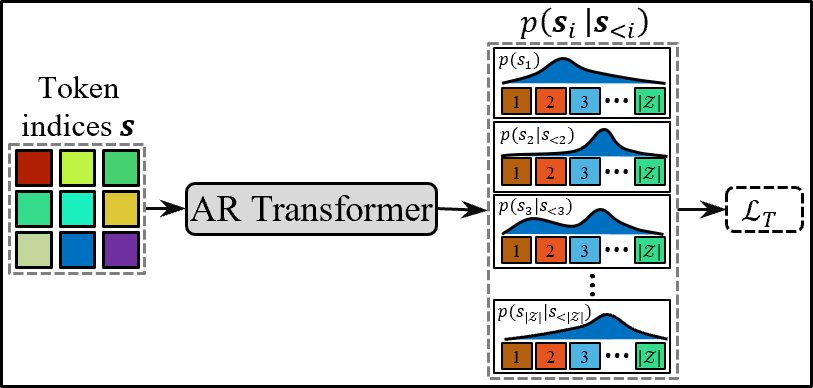}\\{\vspace{-0.13cm}\footnotesize (b) AR transformer model}
        \label{fig:method1}
    \end{subfigure}
    \vspace{-0.3cm}\caption{Overview of the training process.}
    \label{fig:method}
    \vspace{-0.2cm}
\end{figure*}

\section{Method}
The designing process of LidarGRIT involves three steps. First, we represent the Lidar point clouds as range images (Section~\ref{subsec:DR}). Next, we tokenise range images using the VQ-VAE encoder and decode them separately to obtain clean range image along with the raydrop noise mask (Section~\ref{subsec:VQ-VAE}). Finally, we capture the token interactions using the AR transformer (Section~\ref{subsec:ART}).
\subsection{Data Representation} \label{subsec:DR}
We employ different transformations to create range images for the KITTI-360 and KITTI-odometry datasets. For KITTI-360 generation, we use spherical projection, wherein each point in Cartesian coordinates $(x,y,z) \in \RR^{3}$ is transformed into its spherical coordinates $(r, \theta, \phi)$ as:\\
\resizebox{\linewidth}{!}{
    $r = \sqrt{x^2 + y^2 + z^2},\theta = atan(y, x), \phi = atan(z, \sqrt{x^2 + y^2})$.
}\\
We then quantise $\theta$ and $\phi$ into H and W bins with equal bin width, where H and W denote the vertical and horizontal angular resolutions of the Lidar sensor. This yields an image of size $\height \times \width$ with each pixel containing the range of its associated point.
Regarding KITTI-odometry, we employ scan unfolding representation due to the sensor's non-linear vertical spacing~\cite{Nakashima2021LearningTD}. We partition the ordered sequence into $\height$ sub-sequences, with each sub-sequence indicating one elevation angle. Throughout the paper, we denote the input range image as $\rangeimage \in \RR^{\height \times \width}$ and ground-truth raydrop mask as $\mask \in \{0,1\}^{\height \times \width}$.  
\subsection{Adapting VQ-VAE} \label{subsec:VQ-VAE}
We use and adapt VQ-VAE~\cite{DBLP:conf/cvpr/EsserRO21} to auto-encode the range image and raydrop mask for three purposes: tokenisation, downsampling and raydrop noise generation. The VQ-VAE model consists of an encoder $\encoder$, a quantiser $\quantize$ and a decoder $\decoder$. VQ-VAE encoder $\encoder$ downsamples the input range image $\rangeimage$ into latent image $\unquantizedcode = \encoder(\rangeimage) \in \RR^{\text{h} \times \text{w} \times \codebookdim}$. Quantiser $\quantize$ generates tokens $\quantizedcode \in \RR^{\text{h} \times \text{w} \times \codebookdim}$ using a learnable codebook $\codebook = \{z_k\}_{k=1}^K \subset \RR^ {\codebookdim}$.
Finally, VQ-VAE decoder $\decoder$ generates a clean range image $\reconcleanimage \in \RR^{\height \times \width}$ and raydrop mask logits $\masklogit \in \RR^{\height \times \width}$ based on the quantised latent image $\quantizedcode$ as $[\reconcleanimage, \masklogit] = \decoder(\quantizedcode)$. The mask logits are then converted to the binary mask $\reconmask \in \{0,1\}^{\height \times \width}$ via sigmoid function and thresholding:
\begin{equation}
	\reconmask=\begin{cases}
	1 & \sigmoid(\masklogit) \geq0.5 \\
	0 & \sigmoid(\masklogit)<0.5 .
	\end{cases}
\end{equation}
The final generated range image can be obtained as $\reconrangeimage = \reconmask \odot \reconcleanimage$.
\par
\subsubsection{Training--Raydrop Loss} 
We separate the training objectives for range image and raydrop mask generation. To encourage VQ-VAE decoder $G$ to approximate the input range image, we use masked absolute error loss:
\begin{equation}
    \mathcal{L}_{\text{rec}}(\encoder, \decoder) = \expec \big[\frac{1}{ \text{H} \times \text{W}} \Vert \mask \odot (\rangeimage - \reconcleanimage) \Vert_{1} \big] 
\label{eq:l_rec}
\end{equation}
This induces the decoder's range image channel to focus only on estimating the range of existing points. On the other hand, the mask channel is enforced to estimate raydrop noise via raydrop loss: 
\begin{equation}
\begin{split}
     \mathcal{L}_{\text{RL}}(\encoder, \decoder) &=  \expec \big[\text{Avg}[\mask \odot \log( \text{sigmoid}(\masklogit)) \\ &+  (1 - \mask) \odot \log(1 - \text{sigmoid}(\masklogit))] \big],
\end{split}
\label{eq:l_RL}
\end{equation}
Where Avg[$\cdot$] function calculates the average of the input across all its elements.
To align the encoder and codebook embeddings, we train VQ-VAE with so-called commitment loss:
\begin{equation}
        \mathcal{L}_{\text{com}}(\encoder, \codebook) = \expec \big[
   \Vert \text{sg}[\unquantizedcode] - \quantizedcode \Vert_2^2 \nonumber \\ + \Vert \text{sg}[\quantizedcode] - \unquantizedcode \Vert_2^2\big].
\label{eq:l_comm}
\end{equation}
The sg[$\cdot$] operator denotes the straight-through gradient estimator, ensuring that the quantisation process remains differentiable.
Total training loss for the adapted VQ-VAE can be calculated as:
\begin{equation}
    \mathcal{L}_{\text{VQ-VAE}} =\mathcal{L}_{\text{rec}}(\encoder, \decoder)  +   \lambda\mathcal{L}_{\text{RL}}(\encoder, \decoder) \\+  \mathcal{L}_{\text{com}}(\encoder, \codebook).
\label{eq: total_loss}
\end{equation}
By tuning $\lambda$, we can establish a trade-off between the realism of range image generation and that of raydrop mask. We set $\lambda=0.1$ in this study. We show the training process of the adapted VQ-VAE in Figure~\ref{fig:method}\textcolor{red}{a}. 
\subsubsection{Training--Geometric Perseverance}
We observed that the VQ-VAE models, primarily designed for high-resolution RGB images, are prone to overfitting when dealing with relatively low-dimensional range images. This often results in less expressive latent codes. To mitigate the issue, we randomly distort the input images with geometric transformations $\trans$  and push the VQ-VAE to reconstruct the distorted image. This encourages the VQ-VAE to prioritise and preserve the input image geometry. In practice, we randomly replace the input range image $\rangeimage$ and raydrop mask $\mask$ with their respective transformed versions $\trans(\rangeimage)$, $\trans(\mask)$ during the calculation of the training losses (Equation~\ref{eq:l_rec} and~\ref{eq:l_RL}). Our choice of geometric transformations includes affine transformation as well as horizontal and vertical flips.

\begin{figure*}[t!]
\centering
\includegraphics[width=\textwidth]{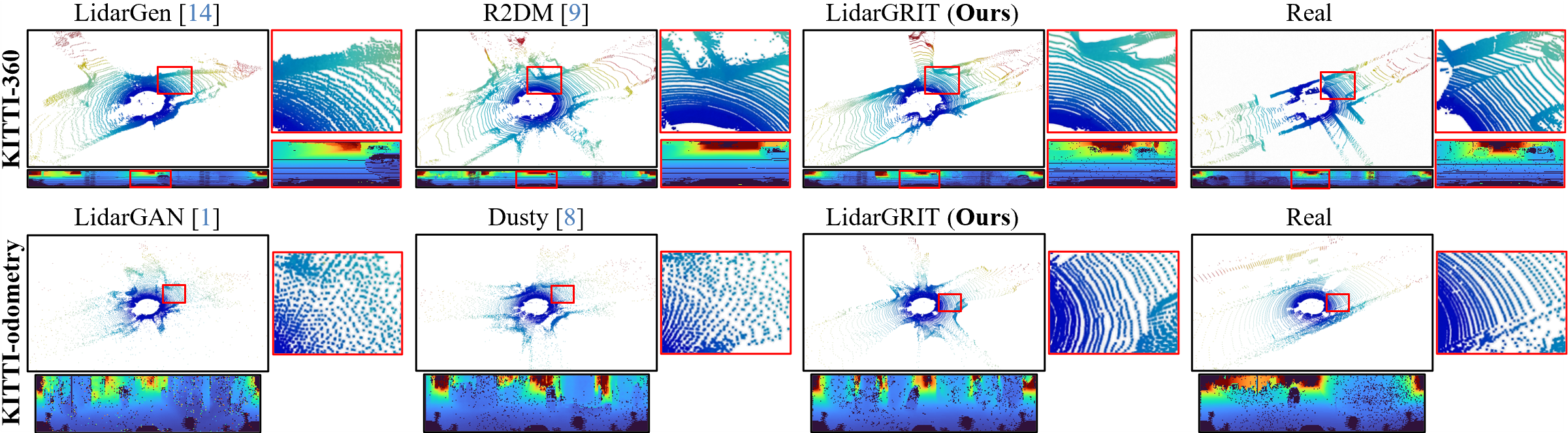}
\caption{Qualitative comparison on KITTI-360~\cite{Liao2022PAMI} and KITTI odometry datasets~\cite{Geiger2013IJRR}.}
\label{fig:qualC}
\end{figure*}

\subsection{Auto-regressive Transformer} \label{subsec:ART}
With trained VQ-VAE, we can encode the range images into token indices $\codeindice \in \big\{ 0, 1, 2, ..., |\codebook|-1\big\}^{\text{h} \times \text{w}}$  and use AR transformers to model interactions between tokens.
 We train the transformer by enforcing auto-regressive modelling. We estimate the likelihood of each token index $\codeindice_{i}$ based on the previous indices $\codeindice_{<i}$ denoted as $p(\codeindice_{i} \vert \codeindice_{<i})$.  The likelihood of the entire sequence can be calculated using chain rule as $p(\codeindice)=\prod_i p(\codeindice_{i} \vert \codeindice_{<i})$. So training objective for the AR transformer can be achieved by negative log-likelihood of $p(\codeindice)$:
\begin{equation}
  \mathcal{L}_{\text{T}} = \expec \left[ -\log p(\codeindice) \right].
\end{equation}
We visualise the training process of the AR transformer in Figure~\ref{fig:method}\textcolor{red}{b}.

\section{Evaluations} 
We detail our experimental settings in Section \ref{subsec:ES}, conduct a comparison to SOTA models in Section \ref{subsec:C-TO-SOTA} and carry out an ablation study in Section \ref{subsec:AS}. 
\subsection{Experimental Settings} \label{subsec:ES}
\textbf{Datasets}
We evaluate our LidarGRIT generations on two datasets: KITTI-360~\cite{Liao2022PAMI} and
KITTI odometry~\cite{Geiger2013}. We follow the Zyrizanov et al~\cite{10.1007/978-3-031-20050-2_2} for partitioning the sequences into train and test splits as well as determining the range image size ($64 \times 1024$). Similarly for KITTI odometry, we follow the setup explained by Nakashima \textit{et al.}~\cite{Nakashima2021LearningTD} for dataset splitting and subsampling the range images to the size of $64 \times 256$.\par
\textbf{Metrics}
We evaluate our generated point clouds against SOTA models using three representations: range image, BEV and point cloud. For range image representation, we employ Sliced Wasserstein Distance (SWD)~\cite{Karras2017ProgressiveGO}. BEV representation is assessed using Maximum-Mean Discrepancy (MMD) and Jensen-Shannon Divergence (JSD) following Zyrizanov \textit{et al.} \cite{10.1007/978-3-031-20050-2_2}. Point cloud evaluation includes Frechet Point Distance (FPD) akin to R2DM~\cite{nakashima2024lidar}, along with minimum-Matching Distance (MD) and JSD similar to Nakashima \textit{et al.}~\cite{Nakashima2021LearningTD}. For calculating MD and JSD metrics, we subsample 512 points from $\height \times \width$ points using farthest point sampling. We create sets of 5k randomly selected generated and real samples for comparison in all metrics.

\par
\begin{table}
\centering
\caption{Quantitative comparison on KITTI-360 dataset.}
\vspace{-0.15cm}
\label{tab:QC-on-kitti-360}
\resizebox{\linewidth}{!}{
\begin{tblr}{
  column{2-5} = {c},
  cell{1}{3} = {c=2}{},
  hline{1,3,7} = {-}{},
  hline{2} = {2-3}{l},
  hline{2} = {3-4}{},
  hline{2} = {4-5}{r}
}
                 & {Image} & {BEV} &               & {Point cloud} \\
Method           & SWD$ \times 10^2 \downarrow$   & MMD$ \times 10^4 \downarrow$         & JSD$ \times 10^2 \downarrow$    & FPD~ $\downarrow$           \\
LidarGen~\cite{10.1007/978-3-031-20050-2_2}         & 33.93              & 2.19               & 5.70          & 43.27                    \\
UltraLiDAR~\cite{xiong2023learning}       & N/A                & 2.23               & 10.52         & N/A                      \\
R2DM~\cite{nakashima2024lidar}            & 20.82              & 4.00               & 4.55          & \textbf{10.84}           \\
LidarGRIT (\textbf{Ours}) & \textbf{10.29}     & \textbf{2.16}      & \textbf{3.93} & 12.54 \\
\end{tblr}
}
\end{table}

\begin{table}
\centering
\caption{Quantitative comparison on KITTI odometry dataset.}
\vspace{-0.15cm}
\label{tab:QC-on-kitti-odemetry}
\resizebox{\linewidth}{!}{
\begin{tblr}{
    column{2-5} = {c},
  cell{1}{3} = {c=3}{},
  hline{1,3,6} = {-}{},
  hline{2} = {2-3}{l},
  hline{2} = {3,4,5}{},
}
                          & {Image}           & { Point cloud} &                              &                  \\
Method                    & SWD$ \times 10^2 \downarrow$ & MD $ \times 10^3 \downarrow$                       & JSD$ \times 10^2 \downarrow$ & FPD $\downarrow$ \\
LidarGAN~\cite{Caccia2018DeepGM}                 & 82.29                        & 6.70                       & 15.98                        & 700              \\
Dusty~\cite{Nakashima2021LearningTD}~                 & 52.81                        & 2.07                       & 5.10                         & 389              \\
LidarGRIT (\textbf{Ours}) & \textbf{15.15}               & \textbf{\textbf{1.65}}     & \textbf{2.06}                & \textbf{116}     
\end{tblr}
}
\vspace{-0.3cm}
\end{table}

\subsection{Comparison to State-of-the-art} \label{subsec:C-TO-SOTA}
We quantitatively compare our LidarGRIT to the SOTA models on KITTI-360 and KITTI odometry generation tasks in Table~\ref{tab:QC-on-kitti-360}~and~\ref{tab:QC-on-kitti-odemetry}, respectively. For KITTI-360 generation, we select the best-performing DMs, R2DM~\cite{nakashima2024lidar} and LidarGen~\cite{10.1007/978-3-031-20050-2_2}, alongside the transformer-based model UltraLiDAR~\cite{xiong2023learning} (notably, the numbers for UltraLiDAR are derived from an unofficial implementation~\cite{ultralidargit} since the public implementation is not available). For KITTI odometry generation, we pick the top-performing GAN-based models Dusty~\cite{Nakashima2021LearningTD} and LidarGAN~\cite{Caccia2018DeepGM}.\\ As shown in the tables, LidarGRIT achieves the best performance on nearly all metrics. Specifically, LidarGRIT obtains significantly superior results on the image-based metric SWD compared with DMs. This can largely be attributed to the more realistic generation of raydrop noise. On the other hand, LidarGRIT substantially outperforms GAN-based models on point cloud representation because of more precise 3D shape modelling via the AR transformer. The results are further evident in the qualitative comparison depicted in Figure~\ref{fig:qualC}. Noteworthy, in the KITTI-360 generation, the adapted VQ-VAE and transformer contain 34.77M and 10.26M parameters respectively, which is
 is comparable to SOTA models such as UltraLiDAR~\cite{xiong2023learning} which contains 40.3M parameters.
\begin{table}
\centering
\caption{Ablation study on KITTI odometry dataset.}
\vspace{-0.15cm}
\label{tab:ablation}
\resizebox{0.9\linewidth}{!}{
\begin{tblr}{
  cell{1}{3} = {c},
  column{1-6} = {c},
  cell{1}{4} = {c=3}{},
  hline{1,3,6} = {-}{},
  hline{2} = {3-4}{l},
  hline{2} = {4,5,6}{},
}
             &              & Image                        & Point cloud   &                              &                  \\
GP           & RL           & SWD$ \times 10^2 \downarrow$ & MD $ \times 10^3 \downarrow$          & JSD$ \times 10^2 \downarrow$ & FPD $\downarrow$ \\
\_           & \_           & 61.73                        & 1.73          & 2.82                         & 212              \\
\_           & $\checkmark$ & 17.68                        & 1.66          & \textbf{1.99}                & 122              \\
$\checkmark$ & $\checkmark$ & \textbf{15.15}               & \textbf{1.65} & 2.06                         & \textbf{116}     
\end{tblr}
}
\vspace{-0.3cm}
\end{table}
\subsection{Ablation Study} \label{subsec:AS}
We evaluate the importance of the two proposed techniques in our VQ-VAE model: raydrop loss (RL) and geometric preservation (GP) in Table~\ref{tab:ablation}. In the baseline scenario (first row), we exclude the raydrop logits channel such that VQ-VAE directly approximates input noisy range images; moreover, we disable the GP technique during training. We introduce the RL in the second row and then incorporate both RL and GP techniques into the baseline in the third row.  As shown, the RL hugely reduces both image-based and point cloud-based metrics due to more accurate raydrop generation. Furthermore, the GP technique improves performance on most metrics up to 14\% by increasing the VQ-VAE generalisability. 
\section{Conclusion} \label{conclusion}
This paper introduced LidarGRIT, a novel Lidar point cloud generative model. The proposed LidarGRIT outperforms SOTA models on KITTI-360 and KITTI odometry datasets. This work also highlighted the importance of the proposed raydrop loss and geometric perseverance in the adapted VQ-VAE model, leading to higher-quality generated samples.

{
    \small
    \bibliographystyle{ieeenat_fullname}
    \bibliography{main}
}


\end{document}